\def\year{2021}\relax
\documentclass[letterpaper]{article} % DO NOT CHANGE THIS
\usepackage{aaai21}  % DO NOT CHANGE THIS
\usepackage{times}  % DO NOT CHANGE THIS
\usepackage{helvet} % DO NOT CHANGE THIS
\usepackage{courier}  % DO NOT CHANGE THIS
\usepackage[hyphens]{url}  % DO NOT CHANGE THIS
\usepackage{graphicx} % DO NOT CHANGE THIS

\usepackage{setspace}
\usepackage{amsmath,amssymb,amsfonts}
\usepackage{algorithm}
\usepackage{algorithmic}
\DeclareMathOperator*{\Max}{max}
\DeclareMathOperator*{\Min}{min}
\usepackage{enumitem}
\usepackage{subcaption}
\usepackage{appendix}

\usepackage{color}
\definecolor{darkred}{RGB}{100,0,0}
\definecolor{darkgreen}{RGB}{0,180,0}
\definecolor{darkblue}{RGB}{0,0,150}
\definecolor{orange}{RGB}{200,100,0}
\usepackage[switch]{lineno} 
\usepackage{multirow}

\usepackage{natbib}  % DO NOT CHANGE THIS AND DO NOT ADD ANY OPTIONS TO IT
\usepackage{caption} % DO NOT CHANGE THIS AND DO NOT ADD ANY OPTIONS TO IT
\nocopyright

\title{Adversarial Context Aware Network Embeddings for Textual Networks}
\author {
        Tony Gracious,\textsuperscript{\rm 1}
        Ambedkar Dukkipati, \textsuperscript{\rm 1}
        
}
\affiliations {
    \textsuperscript{\rm 1} Department of Computer Science and Automation \\
Indian Institute of Science, Bangalore - 560012, INDIA. \\
    
    tonygracious@iisc.ac.in, ambedkar@iisc.ac.in 
}

\begin{document}

\maketitle
%\doublespacing
%\linenumbers
\begin{abstract}
\label{section:abstract}

Representation learning of textual networks poses a significant challenge as it involves capturing amalgamated information from two modalities: (i) underlying network structure, and (ii) node textual attributes. For this, most existing approaches learn embeddings of text and network structure by enforcing embeddings of connected nodes to be similar. Then for achieving a modality fusion they use the similarities between text embedding of a node with the structure embedding of its connected node and vice versa.
%Then for achieving modality fusion they model intra-modal similarities involving networks structure and textual attributes of  nodes in an edge. 
This implies that these approaches require edge information for learning embeddings and they cannot learn embeddings of unseen nodes. In this paper we propose an approach that achieves both modality fusion and the capability to learn embeddings of unseen nodes. The main feature of our model is that it uses an adversarial mechanism between text embedding based discriminator, and structure embedding based generator to learn efficient representations. Then for learning embeddings of unseen nodes, we use the supervision provided by the text embedding based discriminator. In addition this, we propose a novel architecture for learning text embedding that can combine both mutual attention and topological attention mechanism, which give more flexible text embeddings. Through extensive experiments on real-world datasets, we demonstrate that our model makes substantial gains over several state-of-the-art benchmarks. In comparison with previous state-of-the-art, it gives up to 7\% improvement in performance in predicting links among nodes seen in the training and up to 12\% improvement in performance in predicting links involving nodes not seen in training. Further, in the node classification task, it gives up to 2\% improvement in performance.
\end{abstract}

%%%%%%%%%%%%%%%%%%%%%%%%%%%%%%%%%%%%%%%%%%%%%%%%%%%%%%%%%%%%%%%%%%%%%%%%%%%%%%%%%

\section{Introduction}
\label{section:introduction}
With rise in social media and e-commerce websites, there is a huge interest in analyzing these networks for tasks like link prediction, recommendation, community detection, etc. Traditionally, this is done by learning finite-dimensional vector embeddings/representations \cite{deepwalkPerozzi:2014, Grover:2016:node2vec, Tang:2015:LINE} for nodes in these networks and then used it for downstream tasks. One of the challenges is that the quality of these learned representation decreases if the network has many missing links. This affects its performance in downstream tasks. This can be addressed by using attribute similarity of nodes as connected usually have similar attributes. For example, in citation networks, papers on related works will cite each other, and in social media, people with similar interest follow each other. In real-world graphs, nodes of these networks themselves contain rich textual information (e.g., abstract of an article in citation networks \cite{McCallum:2000:Cora,  tangartminter2008, Leskovec:2005:hepth}) as attributes. So, we need techniques which can exploit this textual information while learning node embeddings. The representation learning of textual networks deals with this problem.

Recent methods for representation learning of textual networks involves learning two embeddings, one for the structure information ($z^s$), and the other for the textual information ($z^t$). The embeddings are learned to be similar for nodes that are connected by an edge. The most challenging task is to learn the combined text and structure embeddings, all the previous approaches \cite{CENE2016, tu-etal-2017-cane, Zhang:2018:DMTE, shen-etal-2018-improved, xu-etal-2019-deep, GANE2019} use a joint learning framework by defining a loss function that models the inter-modal similarities between structure and textual information between nodes connected by an edge, in addition to the intra-modal similarities. For example, consider the nodes $v_i$ and $v_j$ with their embeddings  $[z^s_i, z^t_i]$ and $[z^s_j, z^t_j]$. The similarity between embeddings $z^s_i$ and $z^s_j$ is used for modelling intra-model similarity in structure information, on the other hand the similarity between $z^t_i$ and $z^t_j$ is used for intra-model similarity in text information. For inter-model similarity, the similarity between $z^s_i$ and $z^t_j$ is used for modelling the similarity between structure and text, and vice versa. All these similarities are modelled using skip-gram loss function \cite{word2vec2013}. 

The main disadvantage of these models is that they dependent on edge labels for embedding learning. This will make them unable to learn embeddings of nodes which are not present during the training stage. The only way they can be modelled to learn unseen nodes embeddings is by a mapper function between textual information and structure embeddings on seen nodes and apply it to unseen nodes for getting structure embeddings. This can result in a poor performance in downstream tasks involving unseen nodes as the mapping function cannot fully capture the structural information in the nodes. Recenlty, this issue has been addressed by using variational autoencoder framework on the structure and text embeddings~\cite{wang2019improving}. Although it has achieved better performance than the mapper function-based models, the disadvantage of autoencoder framework is that it limits the information learned in the structure embeddings as it is used for predicting the text features by the decoder.

In this paper, we propose an adversarial model where the generator learns the structure embeddings and between text embedding based discriminator and structure embeddings based generator.  For generator, we use the supervision from edge-connectivity and text embedding similarity to learn the structure embeddings. For discriminator model, text embeddings are made dissimilar for node pair generated by the generator and similar for the node pairs from the graph. This training will make the text similarity from the discriminator to approximate the actual similarity in the network. Through this framework we establish that this model efficiently amalgamate or fuse information from both text and graph as both text and structure embeddings use information from both modality for learning. In addition to this, our proposed adversarial approach can be extended for embedding learning of unseen nodes in the training dataset. This is achieved by directly using discriminator based text-similarity as supervision in a post-training stage. This will help in efficiently learning unseen structure embeddings as it does not restrict the embedding learning by using it to predict the text features like in VHE \cite{wang2019improving}.

The performance of the model depends upon how well we can exploit the unstructured textual information, so we need a powerful discriminator. To achieve this, we use context-aware embeddings, where a node has different text embedding for each of its edges. We address this problem by proposing a novel technique by combining two context-aware attention mechanism. The first is based on mutual attention \cite{tu-etal-2017-cane} between word embeddings in text across a pair of nodes.  The other is a topological attention mechanism. This uses structure embeddings of a node pairs to attend over text to learn a topology-aware text embedding. It can reduce the adverse effects of trying to make text embeddings similar where the textual information of connected nodes need not match. Because, this model has better representation capacity as it learns similarity through topological and mutual attention.

 The following are the main contributions of this paper. \textbf{(1)} An adversarial technique for attributed network representation learning. Here, in addition to the supervision from training data, a discriminator using text embeddings is used to give supervision to structure embeddings. \textbf{(2)} A novel text embedding learning technique which uses both mutual and topological attention. \textbf{(3)} Extensive comparative study on downstream tasks of link prediction and node classification. \textbf{(4)} Experiments on link prediction on unseen nodes.

\iffalse
We have evaluated our proposed method on three datasets Cora, Zhihu, and Hepth for link prediction. We observed that our model performs better than state-of-the-art methods in almost all settings in all three datasets. The performance of our model is especially high in low data regime. In Zhihu dataset, our model show a performance improvement of $6.5\%$ over the previous state-of-the-art in the lowest supervision setting. A similar observation was made on the node classification task on Cora dataset, where our adversarial technique achieve state-of-the-art performance. As we mentioned earlier, the main advantage of this model is its ability to the care of representation learning in unseen nodes. We evaluated the quality of these embeddings in link prediction task for edges involving unseen nodes, and ACNE achieves state-of-the-art performance for all settings in all three datasets. On Zhihu dataset, it gave an impressive improvement of  $13\%$ improvement over previous methods in the low-data regime. 
\fi 
\iffalse
\begin{figure}
  \centering
  \includegraphics[width=8cm, height=5cm]{figures/Training.pdf}
   \caption{An overview of the ACNE framework. For a node $v_i$, it samples $v_k$ from  $P_{true}(. | v_i )$  and $v_j$ from $G(. | v_i; Z^s )$. Discriminator is trained to predict 1 for true edge and 0 for generator sampled edge. Generator is trained to approximate ground truth connectivity distribution using the rewards from Discriminator.}
  \label{fig:adversarialmodel}
\end{figure}
\fi 
%%%%%%%%%%%%%%%%%%%%%%%%%%%%%%%%%%%%%%%%%%%%%%%%%%%%%%%%%%%%%%%%%%%%%%%%%%%
\section{Proposed Adversarial Method}
%\subsection{Notation}
Let $\mathbb{G}=(V, E, T) $ represent a network with nodes $V=\{v_1,  \ldots , v_{|V|}\}$, edges $E=\{e_1, \ldots, e_{|E|}\}$, and textual content per node $T=\{t_1,  \ldots ,t_{|V|}\}$. Each textual content, $t_i$, can be reprsented as sequence of word embeddings $t_i = \left<  w_i^1, \ldots, w_i^L \right>$ of length $L$, here $w_i^j \in \mathbb{R}^{d}$ represents embedding of $j^{th}$ word of textual information in $i^{th}$ node. The goal is to learn embedding  $z_i \in \mathbb{R}^{2d}$, for each $v_i \in V$.
%%%%%%%%%%%%%%%%%%%%%%%%%%%%%%%%%%%%%%%%%%%%%%%%%%%%%%%%%%%%%%%%%%%%%%%%%%%

\subsection{Adversarial Technique}
Given a textual network $\mathbb{G}=(V, E, T) $, the node embeddings 
$Z = [z_{1},\ldots,z_{|v|}]$ are nothing but amalgamation of (i) embeddings learned from the textual content $Z^t =[z_1^t, \ldots, z_{|V|}^t] \in \mathbb{R}^{|V| \times d}$ and, (ii) embeddings learned from the structural information $Z^s =[z_1^s, \ldots, z_{|V|}^s] \in \mathbb{R}^{|V| \times d} $. The objective involve two components, a generator and a discriminator. Here generator $G(. | v_i; Z^s )$ tries to approximate the true connectivity distribution $P_{true}(. | v_i )$ using the structure embedding $Z^s$. And, discriminator $D(v_j, v_i;  Z^t)$, tries to predict the probability of a link between the node pairs $(v_i, v_j)$ using the text embedding $Z^t$. %Figure \ref{fig:adversarialmodel} illustrates the overall structure of this adversarial training framework.
The adversarial objective for learning embeddings is, 
\begin{align} \label{eq:advganloss}
 \Min\limits_{Z^s} \Max\limits_{Z^t} & \mathcal{L}_{adv}  (Z^s, Z^t) = \sum_{v_i \in V} \nonumber \\ &  \mathbb{E}_{v_j \sim P_{true}(. | v_i ) }    \left[ \log( D(v_i, v_j; Z^t )) \right]  \nonumber \\
 & +\mathbb{E}_{v_j \sim G(. | v_i; Z^s ) } \left[ \log( 1- D(v_j, v_i; Z^t )) \right] \enspace ,
\end{align}
here embeddings $Z^t$ and $Z^s$ are learned by using adversarial game between the discriminator and generator components. We can see from \eqref{eq:advganloss} that the structure embeddings $Z^s$ use rewards from the discriminator $(\left[ \log( 1- D(v_j, v_i; Z^t ))  \right])$. Since text-similarity is different for different edges, the rewards from discriminator will act as an edge weight for structure embeddings. Further to efficiently capture the fusion of structural and text information, we add supervised information from training data using a cross entropy loss function on the generator. This will improve the similarity of nodes which are similar in textual and structure components. Then the final objective can be written as,

\begin{align} \label{eq:ganloss}
\Min\limits_{Z^s} \Max\limits_{Z^t} & \mathcal{L}(Z^s, Z^t) = \mathcal{L}_{adv}(Z^s, Z^t) \nonumber  \\ & + \eta\mathbb{E}_{v_j \sim P_{true}(. | v_i ) }  \log G(v_j | v_i; Z^s ) \enspace ,
\end{align}
 where $\eta$ is a hyperparameter in the range $[0,1]$ is set to a low value while training in sparse settings. Hence the generator is trained using rewards from the discriminator and supervision from structure of the graph $P_{true}(. | v_i )  $. We will drop $Z^t$ from $D(v_j, v_i; Z^t )$ and $Z^s$ from $G(. | v_i; Z^s )$ when discussing training of these components. In the following section discriminator is denoted as  $D(v_j, v_i)$ and generator is denoted as $G(. | v_i)$.
\subsection{Discriminator} \label{sec: disc}
The role of the discriminator is to predict the presence of links between given nodes. Here, given a node pair $(v_i, v_j)$, the discriminator uses their textual encodings $z_i^t$ and $z_j^t$, respectively, to predict the probability of the edge between them using,
\begin{align} \label{eq:disc_loss}
    D(v_i, v_j) = \sigma( (z_j^t)^T z_i^t) = \frac{1}{1 + \exp{-(z_j^t)^T z_i^t}} \enspace ,
\end{align}
 This model is trained to give a high probability for the edges in the network and low probability for the edges sampled from the generator model. Here generator will act as an adversary to discriminator since it is also trained on discriminator rewards. For each edge in the network, we generate $K$ edges using the generative model and use them as negative samples for training. More details of the architecture of discriminator is given in Section \ref{sec:cne}.

%%%%%%%%%%%%%%%%%%%%%%%%%%%%%%%%%%%%%%%%%%%%%%%%%%%%%%%%%%%%%%%%%%%%%%%%

\subsection{Generator} \label{sec:gen}
The goal of the generator model is to generate nodes that are likely neighbours of the given node. This model is trained with two losses that are represented by the second and third terms in \eqref{eq:ganloss}. The first loss tries to minimize the log probability of discriminator predicting no-edge between a node and its generator sampled neighbour. The second loss tries to make the embeddings of nodes connected by an edge similar. The learning is done by using the gradients from both of these losses. In first loss term, the sampling operation of $v_j$ from the generator in \eqref{eq:ganloss} is discrete, so we incorporate policy gradient for updation \cite{graphgan2018}. The final gradient update from both loss can be written as,
\begin{align} \label{eq:Genupdation}
    \nabla_{Z^s} & \mathcal{L}(Z^s, Z^t)  = 
     \sum_{v_i \in V} \mathbb{E}_{v_j \sim G(. | v_i ) } \nonumber \\ & \nabla_{Z^s} \left[ \log  G(v_j | v_i )  \log( 1- D(v_j, v_i )) \right] \nonumber \\
    & + \eta\mathbb{E}_{v_j \sim P_{true}(. | v_i ) } \nabla_{Z^s} \left[ \log  G(v_j | v_i ) \right] \enspace .
\end{align}
Here the generator outputs the probability of a link between two nodes. This is done using a softmax function over structure embedding similarities of a node to all the other nodes. This is
\begin{align*}
    G(v_j | v_i) = \frac{\exp{ ( (z_j^s)^T z_i^s ) }}{\sum_{j \neq i} \exp{ ((z_j^s)^T z_i^s) }} \enspace .
\end{align*}
Since performing gradient descent operation on $\log  G(. | v_i )$ function is computationally expensive, we use negative sampling \cite{word2vec2013} based approximation, 
\begin{align} \label{eq: Genapprox}
    \log  G(v_j | v_i ) \approx & \log \sigma((z_j^s)^T z_i^s)  \nonumber \\ & +\sum_k^K\mathbb{E}_{v_k \sim \delta_{v_k}^{3/4} } \log \sigma (-1 (z_k^s)^T z_i^s) \enspace ,
\end{align}
where $K$ is the number of negative samples used per edge, and $\delta_{v_k}$ is the degree of the node $v_k$. 

%%%%%%%%%%%%%%%%%%%%%%%%%%%%%%%%%%%%%%%%%%%%%%%%%%%%%%%%%%%%%%%%%%%%%%%%%%%%
\section{ACNE: Proposed Model and the Training} \label{sec:model}
We use context-aware embeddings for learning textual embeddings $Z^t$. So for nodes $v_i$ and $v_j$ in edge $(v_i, v_j)$, context aware text embeddings are $z_{i|j}^t \in \mathbb{R}^d$ and $z_{i|j}^t \in \mathbb{R}^d$, respectively. We concatenate this with structural embedding from a lookup table. Then the context aware embedding for node $v_i$ is $z_{i|j} = [z_i^s, z_{i|j}^t ]$ and for $v_j$ is  $z_{j|i} = [z_j^s, z_{j|i}^t ]$. The final node embeddings are learned by finding the expected value of context-aware embeddings with respect to generator distribution as shown below,
\iffalse
finding the mean of all the context-aware of embedding of node $v_i$ with its edge neighbours.
\begin{align} \label{eq:finalembedding}
z_i = \frac{1}{\delta_{v_i} }\sum_j z_{i | j} \enspace,   
\end{align}
where $\delta_{v_i}$ is the degree of the node $v_i$.
\fi 
\begin{align} \label{eq:finalembedding}
z_i =  \mathbb{E}_{v_j \sim G(. | v_i ) } z_{i | j} \enspace .  
\end{align}

%%%%%%%%%%%%%%%%%%%%%%%%%%%%%%%%%%%%%%%%%%%%%%%%%%%%%%%%%%%%%%%%%%%%%%%%%
\subsection{Context Aware Network Embedding} \label{sec:cne}
Given an edge $(v_i,v_i)$, we use the discriminator model $D(v_i, v_j) = \sigma( (z_{i|j}^t)^T z_{j|i}^t ) $ for learning the context aware embeddings  $z_{i | j}^t$ of node $v_i$ and $z_{j | i}^t$ of node $v_j$. For getting these embeddings, we use two context-aware attention mechanism: (i)  mutual attention mechanism, and (ii) topological attention. The final context-aware embedding can be represented as
\begin{align} \label{eq:contextawareembed}
 z_{i| j }^t = \lambda_{c1} z_{i | j }^{c1} + \lambda_{c2} z_{i | j }^{c2}  \enspace , z_{j |i }^t = \lambda_{c1} z_{j | i }^{c1} + \lambda_{c2} z_{j | i }^{c2} 
\end{align}
where $\lambda_{c1}\ $ and  $\ \lambda_{c2}$ are hyperparameters.

\subsubsection{Mutual Attention} \label{sec: MCoattention}
For implementing the mini-batch training, we pad all the text sequence to equal length of $L$, i.e. every text sequence $t \in \mathbb{R}^{L \times d}$. Then we create masking vectors $M_i$ and $M_j \in \mathbb{R}^{L \times 1}$ with zeros corresponding to the padding vectors in respective text sequence. Then for finding content-aware embedding $z_{i|j }^{c1}$ and $z_{j|i}^{c1}$ we apply mutual attention.% as shown in Figure \ref{fig:model}. 
That involves operations $M  = M_i (M_j)^T , \enspace A  = t_i (t_j)^T$ and $\mathcal{A} = \mathrm{tanh}(A \odot M )$.
\iffalse
\begin{align} \label{eq:affinity}
    M & = M_i (M_j)^T \enspace , \enspace 
    A  = t_i (t_j)^T \text{ and}  \nonumber \\
    \mathcal{A} & = \mathrm{tanh}(A \odot M ).
\end{align}
\fi
Here $A \in \mathbb{R}^{L \times L}$ shows the affinity between words in both text sequences and $\odot$ is element-wise multiplication. Then we find the word importance score using row-wise and column-wise mean pooling as
\begin{align} \label{eq:imp-scores}
    c_{i|j} &= \frac{\sum_{n=1}^L \mathcal{A}_{mn}}{\sum  M_j} \:, \text{ and}\:\:
    c_{j | i} =  {\left(\frac{\sum_{m=1}^L \mathcal{A}_{mn}}{\sum M_i} \right)}^ T \:,
\end{align}
where $c_{i|j} \in \mathbb{R}^{L}$ is the importance score vector of words in $t_i$ with respect to $t_j$, and $c_{j |i}$ is similarly defined. For finding the final context-aware importance scores, we normalized the weights of $c_{i | j}$ and $c_{j | i}$ as 
\begin{align}
    \beta_{i | j} &= \frac{ \exp{(c_{i | j} ) } \odot M_i }{\sum(\exp{(c_ {i | j} ) } \odot M_i) } \enspace ,   \nonumber\\
    \beta_{j | i} &= \frac{ \exp{(c_{j | i} ) } \odot M_j }{\sum(\exp{(c_ {j | i} ) } \odot M_j) } \enspace .
\end{align}
The final embeddings are found as shown below found by using $\beta_{i | j}$ and  $\beta_{j | i} \in \mathbb{R}^{L} $ to find the weighted average of $t_{i}$ and $t_{j}$, $z_{i| j}^{c1} =  \beta_{i |j}^T t_i$, and $ z_{ j | i }^{c1} =  \beta_{j |i}^T t_j$.
\iffalse
\begin{align}
z_{i| j}^{c1} &=  \beta_{i |j}^T t_i \enspace , \text{ and } 
z_{ j | i }^{c1} =  \beta_{j |i}^T t_j \enspace .
\end{align}
\fi

%%%%%%%%%%%%%%%%%%%%%%%%%%%%%%%%%%%%%%%%%%%%%%%%%%%%%%%%%%%%%%%%%%%%%%%%%

\subsubsection{Topological Attention} \label{sec:topattention}
For learning context-aware embeddings from topological attention, we use the structural embeddings of node pair in an edge  to attend over the text. Then the embeddings learned will not be affected by the diverse content in the neighbouring nodes, which happens when we use mutual attention between textual content. Give a node pair $v_i$ and $v_j$ with its structural embeddings $z^s_i$ and $z^s_j$, we extract context-aware embedding for node $v_i$ by using structural embeddings to attend over its embedded text sequence $t_i \in \mathbb{R}^{L \times d}$. We extract two features, first using self-interaction between $z^s_i$ and $t_i \in \mathbb{R}^{L \times d}$ as,

\begin{align} \label{eq:top1}
    \gamma_i^s &= \frac{\exp{( \mathrm{tanh}(t_i  z^s_i)) }}{ \sum \exp{ (\mathrm{tanh}(t_i  z^s_i)) } } \enspace , \text{ and }
    z^{self}_i = (\gamma_i^s)^T t_i \enspace .
\end{align}
The second feature using cross-interaction between $z^s_j$ and $t_i \in \mathbb{R}^{L \times d}$ as shown below.
\begin{align} \label{eq:top2}
    \gamma_{i|j}^c = \frac{\exp{ (\mathrm{tanh}(t_i  z^s_j)) }}{ \sum \exp{ (\mathrm{tanh}(t_i  z^s_j) ) } } \enspace , \text{ and }
    z^{cross}_{i|j} = (\gamma_{i|j}^c)^T t_i \enspace .
\end{align}
%Figure \ref{fig:model2} shows the architecture of this model.
The final context-aware embedding $z^{c2}_{i|j} = \lambda_s z^{self}_i  + \lambda_c  z^{cross}_{i|j}$ can be represented as weighted sum of both,  where $\lambda_s\ $ and $\ \lambda_c $ are hyperparameters. Similarly, one can find $z^{c2}_{j|i}$.

%%%%%%%%%%%%%%%%%%%%%%%%%%%%%%%%%%%%%%%%%%%%%%%%%%%%%%%%%%%%%%%%%%%%%%%%%%%%

%%%%%%%%%%%%%%%%%%%%%%%%%%%%%%%%%%%%%%%%%%%%%%%%%%%%%%%%%%%%%%%%%%%%%%%%%%%%%%%%%%%%%%%%%%%%%%%%%%%%
\subsection{Training} \label{sec: training}

\begin{algorithm}[tb]
\caption{Learning ACNE}
\label{alg:algorithm2}
\textbf{Input}: $\mathbb{G} $\\
\textbf{Parameter}: dimension of the embeddings $d$, number of negative samples $K$, number of epochs $\mathsf{E}$ , minibatch size $B$ \\
\textbf{Output}: $Z^s, Z^t$
\begin{algorithmic}[1] %[1] enables line numbers
%\STATE Let $epoch=0$.
\STATE Pre-train  $G(v_j | v_i, Z^s)$
\FOR{$\mathsf{E}$-steps}
\FOR{ $\mathsf{ D}$-steps} 
\STATE Sample $B$ positive samples from the ground truth edges and $BK$ negative samples using $G( . | v, Z^s)$
\STATE Learn $Z^t$ to maximize \eqref{eq:disc_loss} for true edges using the architecture mentioned in Section \ref{sec:cne}
\ENDFOR
\FOR{$\mathsf{G}$-steps}
\STATE Sample $B$ vertices from $V$
\STATE For each vertex $v_i$, sample a node from $G(v_j | v_i, Z^s)$ and $P_{true}(. | v_i )$ 
\STATE Update $Z^s$ using \eqref{eq:Genupdation}
\ENDFOR
\ENDFOR
\STATE Find the final aggregate $Z$ using \eqref{eq:finalembedding}
\STATE \textbf{return} $Z$
\end{algorithmic}
\end{algorithm}

At first, we pretrain the generator using the third component of the loss term \eqref{eq:ganloss}. This component uses the edge connectivity information in the training dataset and make embeddings of nodes similar if they have an edge between them. Then for learning the generator embedding $Z^s$ and discriminator text embedding $Z^t$, we use an alternate optimization strategy with two stages as shown in Algorithm \eqref{alg:algorithm2}. 

At the first stage, we train the discriminator by using context-aware text encoding features learned from Section \ref{sec:cne}. For that, we sample $B$ samples from the ground truth edges, where $B$ is the mini-batch size, and for each true edge, we sample a negative sample by using the generator $G(. | v)$. Then \eqref{eq:disc_loss} is maximized for true edges and minimized for negative samples from the generator. This operation will be repeated for $\mathsf{D}$-steps. While training discriminator, we do not update the node-embeddings used in the topological attention \ref{sec:topattention}.

In the second stage, for training generator, we sample $B$ vertices from the training data. Here, $B$ is the batch-size of the generator. For each vertex in mini-batch, we sample a node from $G(v_j | v_i, Z^s)$ and $P_{true}(. | v_i )$. Then generator embeddings are trained using the update rule mentioned in \eqref{eq:Genupdation}. This operation will be executed for $\mathsf{G}$-steps.

\iffalse
This alternating optimization strategy by keeping either generator or discriminator fixed and improving the other will help in distributing the textual information to structure embedding learning and vice-versa. This will fuse the graph and text modalities information better than the cross-modal similarity based loss function used by previous methods \cite{CENE2016, tu-etal-2017-cane, Zhang:2018:DMTE, shen-etal-2018-improved, xu-etal-2019-deep,GANE2019,wang2019improving}. Another advantage of this strategy is that the negative sampling for training is done adaptively based on the generator distribution compared to previous methods which use a fixed sampling distribution \cite{word2vec2013}.  This will help in building robust discriminator as it trained to predict false for increasing difficult fake edges from the generator as the training progress. 
\fi

We repeat the alternating optimization for $\mathsf{E}$ epochs. Once the training is done,  the final node embeddings are learned by aggregating all the context-aware embedding of a node using \eqref{eq:finalembedding}.

\subsection{Unseen Nodes Embedding Learning} \label{sec: unseennodes}
For learning the embeddings for unseen nodes $\Tilde{V}$ during training, we have a post-training stage where we fix the parameters of the discriminator model and structure embeddings of seen nodes obtained from the training stage. Then optimize the loss $\mathcal{L}([Z^s, \Tilde{Z}^s ], [Z^t, \Tilde{Z}^t])$ with respect to structure embeddings $\Tilde{Z}^s$ of unseen nodes $\Tilde{V}$. Here $\Tilde{Z}^t$ is the  text embeddings of unseen nodes and is kept fixed as the discriminator parameters are kept fixed. The objective function for this is 
\begin{align} \label{eq:unseennodesloss}
    \Min\limits_{\Tilde{Z}^s }    \sum_{v_i \in \Tilde{V} }  \mathbb{E}_{v_j \sim G(. | v_i; [Z^s, \Tilde{Z}^s ]  ) }  \left[ \log( 1- D(v_j, v_i; [Z^t, \Tilde{Z}^t])) \right] \enspace .
\end{align}
Here, learning is done through the supervision provided by the text-embedding based discriminator. For initializing the embeddings of unseen nodes before \eqref{eq:unseennodesloss}, we learn a mapper function $MLP(t_i)$ to map text sequence $t_i$ to node embedding $z^s_i$ of seen nodes $v_i$. The architecture of the function is
\begin{align}
    x^j_i  &= C  w^{j:j+l-1}_i + b \enspace , \nonumber \\
    MLP( t_i)   &= P_2(\phi( P_1( \phi( \mathrm{Max}(x^0_i, \ldots  , x^L_i ) ) ) ) )  \enspace ,
\end{align}
where $C \in \mathbb{R}^{ d \times l \times d}$ is the convolutional filter of window size $l$, $b \in \mathbb{R}^d $ is the bias vector, $x_j^i \in \mathbb{R}^d$ is the ouput of convolutional filter, $\mathrm{Max}$ is a max pooling operation, $\phi$ is the $\mathrm{ReLU}$ non-linearity function, and  $P_1 , P_2 \in \mathbb{R}^{d \times d}$ are the projection matrices.  We minimize $\sum_i^{|V|}\| MLP(t_i) - z^s_i\|^2$ loss function to learn the mapping function parameters. We then use this mapper function to initialize the embeddings $\Tilde{Z}^s$ of unseen nodes using their textual information.We also add this loss function to \eqref{eq:unseennodesloss} as a regularizer. Then the modified objective function is 
\begin{align} \label{eq:unseennodesregularizedloss}
     \Min\limits_{\Tilde{Z}^s }    \sum_{v_i \in \Tilde{V} }  \mathbb{E}_{v_j \sim G(. | v_i; [Z^s, \Tilde{Z}^s ]  ) }  \left[ \log( 1- D(v_j, v_i; [Z^t, \Tilde{Z}^t])) \right] \nonumber \\ + \lambda_r \sum_{v_i \in \Tilde{V} }  \|MLP( t_i) - \Tilde{z}^s_i \|_2^2  \enspace ,
\end{align}
where $\lambda_r$ is a hyperparameter and the parameters of $MLP(t_i)$ are kept fixed during post training stage.
%%%%%%%%%%%%
\section{Experiment Setting}
The learned embeddings are evaluated on downstream tasks of link-prediction and multi-label node classification. 

%%%%%%%%%%%%%%%%%%%%%%%%%%%%%%%%%%%%%%%%%%%%%%%%%%%%%%%%%%%%%%%%%%%%%%%%%%%%%
%%%%%%%%%%%%%%%%%%%%%%%%%%%%%%%%

\begin{table}[!ht] 

\begin{center}
 \begin{tabular}{| c | c| c | c |}
 \hline
 Datasets & Zhihu & Cora  &  Hepth \\
\hline
Vertices  & 10000 & 2277 & 1038 \\
Edges & 43894 & 5214 & 1990 \\
%Max Degree & 2191 & 93 & 24 \\
Avg text length & 190 & 90 & 54  \\
%Max text length & 265 & 410 & 148  \\
labels & NA & 7 & NA \\
\hline
\end{tabular}
\caption{Dataset Statistics}
\label{tab: datasetstatistic}
\end{center}
\end{table}

%%%%%%%%%%%%%%%%%%%%%%%%%%%%%%%%%%%%%%%%%%%%%%%%%%%%%%%%%%%%%%%%%%%%%%%%%%%%%%%%
\subsection{Datasets and Baselines}
We evaluated our model on datasets Cora \cite{McCallum:2000:Cora}, Zhihu \cite{CENE2016}, and Hepth \cite{Leskovec:2005:hepth}. The summary statistics of these datasets are shown in \ref{tab: datasetstatistic} and additional information is provided in the supplementary material.
\iffalse
In this Cora and Hepth are citation datasets, and Zhihu is a network created from Chinese Q \& A social media. The summary statistics of these datasets are shown in the supplementary material. %The datasets contain edges between nodes, textual content in all the nodes and node-labels are available for Cora dataset.
\fi 

To evaluate our model, we compared ACNE's performance on downstream tasks against previous state-of-the-art models. They are, i) Structure based network representations: Deepwalk \cite{deepwalkPerozzi:2014}, Node2vec \cite{Grover:2016:node2vec}, GraphGAN \cite{graphgan2018}, and ii) Content augmented network representations: TADW \cite{Yang:2015:NRL:TADW}, Tri-DNR \cite{Pan:2016:TriDN}, CENE \cite{CENE2016}, CANE \cite{tu-etal-2017-cane}, DMTE \cite{Zhang:2018:DMTE}, WANE \cite{shen-etal-2018-improved}, NEIFA \cite{xu-etal-2019-deep}, GANE \cite{GANE2019}, VHE \cite{wang2019improving}.
\subsection{Evaluation Metrics and Parameter Settings}
We use the AUC metric \cite{auc1982} for evaluating link-prediction task. Since testing data only have ground truth edges, we create fake edges for each of the edge in the test by replacing its source or target node by a randomly chosen node. We then rank the similarity of embeddings of nodes in true edges and fake edges. Then use AUC to calculate the probability that nodes in true edges are similar than nodes in fake edges. The experiment is repeated for 10 times for each training ratio, and the average score is reported in the tables.

For evaluating node-classification task, we use the node embeddings and labels for training a classifier. Then we evaluate the classifier prediction using Macro F1-score measure. We do this experiment for different levels of supervision, and each experiment is repeated for 10 times, and the average score is reported. 

For getting a fair comparison with previous methods, we fix the embedding size of the textual embedding $z^t$ and structure embedding $z^s$ to $100$, same as previous works. The word embedding size is also fixed at $100$ dimensions. We keep the number of negative samples ($K$) for the discriminator and generator training to $1$. The number of gradient descent steps per iteration for discriminator optimization $\mathsf{D}$ is set as $2$ and for generator optimization steps $\mathsf{G}$ is set to $1$. All the text is padded to a length of $300$. For all the experiments in link prediction and node classification, our model is trained using a mini-batch size $B$ of $256$. The hyper-parameters $\eta, \lambda_{c1}, \lambda_{c2}, \lambda_s, \lambda_c, \lambda_r$ are tuned in the range $[0, 1]$ during validation. For unseen nodes link prediction, we train the model using a mini-batch size of $32$ and mapper function $MLP(t)$ uses a mini-batch size of $512$. Our model is implemented in PyTorch \cite{pytorch2019} and use its Adam optimizer \cite{DBLP:journals/corr/KingmaB14} for training the model. For all the experiments, we kept the learning rate fixed at $0.001$. For training the baseline models, we used a joint learning framework used by CANE \cite{tu-etal-2017-cane}. Experimental settings and hyper-parameter details of ACNE and its baselines for various experiments are discussed in the supplementary material.

%%%%%%%%%%%%%%%%%%%%%%%%%%%%%%%%%%%%%%%%%%%%%%%%%%%%%%%%%%%%%%%%%%%%%%%%%%%%%%%%%%%%%%%%%%%%%%%%%%%%%%%%%%%%%%%%%%%%%%%%%%%%%%

%%%%%%%%%%%%%%%%%%%%%%%%%%%%%%%%%%%%%%%%%%%%%%%%%%%%%%%%%%%%%%%%%%%%%%%%%%%%%%%%%%%%%%%%%%%%%%%%%%%%%%%%%%%%%%%%%%%%%%%%%%%%%%

\begin{table*}[!ht]
\setlength{\tabcolsep}{4pt}
\begin{center}
 \begin{tabular}{l| c c c c c | c c c c c |  c c c c c | }
 \hline 
 \textbf{Data}    & \multicolumn{5}{c|}{\textbf{cora}} & \multicolumn{5}{c|}{ \textbf{Hepth}} & \multicolumn{5}{c|}{\textbf{Zhihu}}   \\
 \hline
 $\%$ of edges & 15$\%$ & 35$\%$ & 55$\%$ & 75$\%$ & 95$\%$ & 15$\%$ & 35$\%$ & 55$\%$ & 75$\%$ & 95$\%$ & 15$\%$ & 35$\%$ & 55$\%$ & 75$\%$ & 95$\%$  \\  
\hline
Deep Walk$^a$  & 56.0 & 70.2 & 80.1  & 85.3  & 70.3 & 55.2 & 70.0 & 81.3 & 87.6 & 88.0 & 56.6 & 60.1 & 61.8 & 63.3 & 67.8 \\
%LINE$^b$  & 55.0  &  66.4  & 77.6  &  85.6  & 89.3 & 53.7 & 66.5 & 78.5 & 87.5 & 87.6 & 52.3 & 59.9 & 64.3 & 67.7 & 71.1\\
node2vec$^b$ & 55.9  &  90.2  & 78.7 & 85.9 & 88.2 & 57.1 & 69.9 & 84.3 & 88.4 & 89.2 & 54.2 & 57.4 & 58.7 & 66.2 & 68.5\\
GraphGAN$^c$ & 59.3  & 71.3  & 82.3  & 87.5  & 91.1 & 60.6 & 71.9  & 84.8  & 88.9  & 91.1 & 57.1 & 60.2 & 64.4 & 68.2 & 71.3\\
\hline
TADW$^d$  & 86.6  & 90.2  & 90.0 & 91.0  & 92.7 & 87  & 91.8 & 91.1 & 93.5 & 91.7 & 52.3 & 55.6 & 60.8 & 65.2 & 69.0\\
TriDNR$^e$  & 85.9  &  90.5 & 91.3  & 93.0 & 93.7 & \_ & \_ & \_ & \_ & \_ & 53.8 & 57.9 &  63.0 & 66.0 & 70.3\\
CENE$^f$ & 72.1 & 84.6  &  89.4  &  93.9 & 95.9 & 86.2 & 89.8 & 92.3 & 93.2 & 93.2  & 56.2 & 60.3 & 66.3 & 70.2 & 73.8\\
CANE$^g$  & 86.8 & 92.2  &  94.6  & 95.6 & 97.7 & 90.2 & 92.0 & 94.2 & 95.4 & 96.3 & 56.8 & 62.9 & 68.9 & 71.4 & 75.4  \\
WANE$^h$  & 91.7 &  94.1  & 96.2 & 97.5  & 99.1 & 93.7 & 95.7 & 97.5 & 97.7 & 98.7 & 58.7 & 68.3 & 74.9 & 79.7 &  82.6\\      
DMTE$^i$  & 91.3  & 93.7  & 96.0  & 97.4  & 98.8 & \_ & \_ & \_ & \_ & \_  & 58.4 & 67.5 & 74.0 & 78.7 & 82.2\\
NEIFA$^j$  & 89.0  & 95.3  &  97.1  & 97.6  &  99.2 & 91.7 & 95.9 & 97.4 & 98.0 & 99.1 & 68.9 & 78.3 & 84.5 & 88.2 & 90.1\\
GANE$^k$ & 94.0  & 97.2 & 98.0  & 98.8  & 99.3 & 93.8 & 97.3 & 98.1 & 98.4 & 98.9 & 64.6 & 72.8 & 79.1 & 81.8 & 84.3 \\
VHE$^l$  & 94.4  & \textbf{97.6}  & 98.3  & \textbf{99.0} & 99.4 & 94.1 & 97.5 & 98.3 & 98.8 & \textbf{99.4} & 66.8 &  74.1 &  81.6 &  84.7 &  86.4\\
\hline 
CNE-mu (baseline) & 94.2  & 96.4  & 98.1  & 98.7  & 99.2 & 95.1 & 97.0 & 98.1  & 98.8  & 99.2  & 67.3 & 76.5 & 79.5 & 82.7 & 85.9\\  
CNE-top (baseline) & 91.4  &  94.8  & 96.3  & 97.7  & 98.8 & 89.8  & 94.7& 96.8  & 97.6 & 99.1  & 71.7 & 79.3 & 84.3 & 87.4 & 89.8\\  
CNE (baseline) & 94.4 & 97.0  &  98.3  & 98.8 & 99.3 & 95.5 & 97.1  & 98.2  & 98.8  &  99.2 & 72.1 & 79.4 & 84.4 &  89.0 & 93.0 \\  
\hline
ACNE-mu (baseline) & 95.0& 97.3 & 98.3 & 99.0 & 99.4 & 96.0 & 97.2 & 98.2 & 98.9 & 99.3 & 69. & 77.9 & 81.2 & 85.3 & 88.3 \\
ACNE (ours) & \textbf{95.8}  & \textbf{97.6} & \textbf{98.5}   & \textbf{99.0}   & \textbf{99.5}  & \textbf{96.3}  & \textbf{97.9}  & \textbf{98.5}  & \textbf{99.0}  & \textbf{99.4} & \textbf{73.4} & \textbf{82.4} & \textbf{88.6} & \textbf{91.1} & \textbf{93.2} \\  

\hline
\end{tabular}
\caption{AUC scores for link prediction in Cora, Hepth and Zhihu datasets. Here a higher value of AUC indicates better performance. \textbf{Citations: }$^a$\cite{deepwalkPerozzi:2014}, $^b$\cite{Grover:2016:node2vec} , $^c$\cite{graphgan2018}, $^d$\cite{Yang:2015:NRL:TADW}, $^e$\cite{Pan:2016:TriDN}, $^f$\cite{CENE2016}, $^g$\cite{tu-etal-2017-cane}, $^h$\cite{shen-etal-2018-improved}, $^i$\cite{Zhang:2018:DMTE}, $^j$\cite{xu-etal-2019-deep}, $^k$\cite{GANE2019}, $^l$\cite{wang2019improving} }
\label{tab:linkpredcitation}
\end{center}

\end{table*}

\section{Results}

%%%%%%%%%%%%%%%%%%%%%%%%%%%%%%%%%%%%%%%%%%%%%%%%%%%%%%%%%%%%%%%%%%%%%%%%%%%%%%%%%%%%%%%%%%%%%%%%%%%%%%%%%%%%%%%%%%%%%%%%%%%%

\subsection{Link Prediction}
The goal of this task is to predict links between nodes using the inner product similarity of their respective node embeddings. For comparing the performance of our model against previous state-of-the-art models, we experiment in different settings by varying percentage of training edges used from $15\%$ to $95\%$ at steps of 20 and rest is used for testing. We use datasets Cora, Hepth, and Zhihu datasets for these experiments and model performance is shown in Table \ref{tab:linkpredcitation}. Here, the models that uses textual information along with structural information considerably outperforms the models that only use structural information. This is evident from the low-performance of models like DeepWalk, Node2vec, and GraphGAN compared to models that use textual information for embedding learning. We can see that our model ACNE perform better than baseline model CNE, which does not use adversarial training strategy.
\iffalse
This because in adversarial training, we use an attribute-based discriminator for providing supervision for edges sampled using structure embeddings based generator. Through this model, we can give different weight-age to edges based on the similarity of their text embedding. This differential treatment of edges will help to identify truly similar nodes. This is important as it is often the case that edges between nodes don't mean that they have high similarity. This is common in social media networks and citation networks where nodes from the different communities having links.
Our model can identify these edges based on their text-similarity and give less weight-age to them.\fi 
The performance of our model in the low-data regime of $15\%$ training data is much higher than state-of-the-art models. For Zhihu dataset in low data regime, we are getting an improvement of around $6\%$. This is due to the combined effect adversarial and context-aware embedding learning we apply in our model. This helped our model achieves state-of-the-art performance in almost all settings. The results shown for previous models except for GraphGAN \cite{graphgan2018} are taken from their respective papers. For GraphGAN, we used an open source implementation \footnote{https://github.com/liutongyang/GraphGAN-pytorch} with discriminator and generator embeddings initialized with Node2vec pretraining.
%%%%%%%%%%%%%%%%%%%%%%%%%%%%%%%%%%%%%%%%%%%%%%%%%%%%%%%%%%%%%%%%%%%%%%%%%%%%%%%%%%%%%%%%%%%%%%%%%%%%%
\iffalse
\begin{figure*}
\centering
\begin{subfigure}{.32\textwidth}
  \centering
  \includegraphics[width=5cm, height=5cm]{figures/link_pred_Cora.pdf}
   \caption{Link Prediction on Cora}
  \label{fig:ablationlinkpredcora}
\end{subfigure}
\begin{subfigure}{.32\textwidth}
  \centering
  \includegraphics[width=5cm, height=5cm]{figures/link_pred_zhihu.pdf}
   \caption{Link Prediction on Zhihu}
  \label{fig:ablationlinkpredzhihu}
\end{subfigure}
\begin{subfigure}{.32\textwidth}
  \centering
  \includegraphics[width=5cm, height=5cm]{figures/link_pred_HepTH.pdf}
   \caption{Link Prediction on Hepth}
  \label{fig:ablationlinkpredhepth}
\end{subfigure}
    \caption{Ablation Study on link prediction. We can observe that the adversarial techniques ACNE and ACNE-mu outperforms their non-adversarial counterparts ACNE and ACNE-mu, respectively. Also, our model ACNE outperforms all of them.}
    \label{fig:ablationlinkpred}
\end{figure*}
\fi
%%%%%%%%%%%%%%%%%%%%%%%%%%%%%%%%%%%%%%%%%%%%%%%%%%%%%%%%%%%%%%%%%%%%%%%%%%%%%%%%%%%%%%%%%%%%%%%%%%%%%%%%%%%%%%%%%%%%%

\begin{table*}[!ht]
\setlength{\tabcolsep}{5pt} 
\begin{center}
 \begin{tabular}{|c| c|c|c|c|c | c|c|c|c|c | c|c|c|c|c |}
 \hline 
 Data    & \multicolumn{5}{c|}{Cora} & \multicolumn{5}{c|}{Hepth}  & \multicolumn{5}{c|}{Zhihu}  \\
 \hline
 $\%$ of Nodes & 15$\%$ & 35$\%$ & 55$\%$ & 75$\%$ & 95$\%$ & 15$\%$ & 35$\%$ & 55$\%$ & 75$\%$ & 95$\%$ & 15$\%$ & 35$\%$ & 55$\%$ & 75$\%$ & 95$\%$  \\ 
 \hline
 CANE $^a$ & 83.1 & 86.8 & 90.4 & 93.9 &  95.2 &  83.8 & 88.0 & 91.0 & 93.7 & 95.0 & 56.0 & 61.5 & 66.9 & 73.5 & 76.3 \\
WANE  $^b$ & 87.0 & 88.8 &  92.5 & 95.4 & 95.7 & 86.6 & 88.4$^*$ & 91.0 & 93.7 & 95.0 & 57.8 & 65.2 & 70.8 & 76.5 & 80.2 \\ 
VHE  $^c$ & 89.9 & 92.4 & 95.0 & 96.9 & 97.4 & 90.2 & 92.6 & 94.7 & 96.6 & 97.7 & 63.2 & 75.6 & 78.0 & 81.3 & 82.7 \\
\hline
CNE (baseline) &  80.2 & 90.3 & 94.3 & 97.1 &  98.2 & 84.2 & 86.9 & 93.3 & 96.3 & 97.8 & 65.3 & 74.3 & 77.4 & 80.3 & 81.1 \\
\hline 
ACNE (ours ) & \textbf{92.3} & \textbf{93.8} &  \textbf{95.5} & \textbf{97.2} & \textbf{98.9} & \textbf{91.0} & \textbf{94.1} & \textbf{94.9} & \textbf{96.6} & \textbf{98.2} & \textbf{71.9} & \textbf{78.1} & \textbf{81.6} & \textbf{83.4} & \textbf{84.3}\\ 
%AMCNE (ours) & 88.9$^*$ & 91.3$^*$ & 93.5$^*$ & 96.0$^*$ & \textbf{99} & \textbf{91.2} & \textbf{93.3} & \textbf{95.3} & \textbf{96.6} & \textbf{99.2} & \textbf{68.0} & 72.5$^*$ & 77.1$^*$ & 80.0$^*$ & \textbf{84.1} \\
 
\hline
\end{tabular}
\end{center}
\caption{AUC scores for link prediction in Cora, Hepth and Zhihu datasets for unseen nodes. Here a higher value of AUC score indicates better performance. \textbf{Citations: }$^a$\cite{tu-etal-2017-cane},$^b$\cite{shen-etal-2018-improved}, $^c$\cite{wang2019improving} }
\label{tab:linkpredunseen}
\end{table*}
%%%%%%%%%%%%%%%%%%%%%%%%%%%%%%%%%%%%%%%%%%%%%%%%%%%%%%%%%%%%%%%%%%%%%%%%%%%%
%

%%%%%%%%%%%%%%%%%%%%%%%%%%%%%%%%%%%%%%%%%%%%%%%%%%%%%%%%%%%%%%%%%%%%%%%%%%%%%%%%%%%%%%%%%%%%%%%%%%%%%%%%%%%%%%%%%%%%%%%%%%%%%%

\paragraph{Ablation Studies.} For showing the effectiveness of the architecture of model ACNE, we have created 3 variants of the model. 1) CNE: a non-adversarial version of ACNE. 2) CNE-top: this model uses only topological attention for text embedding. 3) CNE-mu: this model uses only mutual attention for text embedding. The results for these models for Cora, Hepth, and Zhihu are shown in Table \ref{tab:linkpredcitation}. We can observe that CNE give better performance over  CNE-mu and CNE-top models. In Zhihu dataset, CNE and CNE-top perform better than CNE-mu. This is because Zhihu is a social media website, and the textual content of two connected people in social media need not match because people with different views can follow each other on social media. This resulted in poor performance of text-matching base models like CANE, WANE, GANE, VHE and CNE-mu. Further, we can see that CNE-mu performs well in citation networks Cora and Hepth. This is because connected nodes have similar textual information in this type of networks. Whereas CNE-top does not perform well in the sparse setting of these datasets as the performance is much less that of CNE-mu. This is because CNE-mu model has less dependence on structure embedding than CNE-top as the sparse setting reduce the quality of node embeddings. Our model CNE combines both mutual and topological attention into a joint learning framework for learning better representations. This helped in getting better performance than the models using a single type of context-aware attention mechanism. 

For showing the effectiveness of adversarial training, we compared adversarial models ACNE and ACNE-mu with their non-adversarial counterparts CNE and CNE-mu, respectively. From Table \ref{tab:linkpredcitation}, we can see that adversarial loss function gives better performance over non-adversarial models with joint loss function used by \cite{tu-etal-2017-cane}. The non-adversarial techniques treat edges with equal weightage, and sampling technique they use is from a fixed distribution proportional to the degree of the nodes. Unlike these, adversarial techniques use differential treatment of edges due to the use of rewards from the discriminator, and they follow an adversarial sampling of nodes for the training discriminator. These help in combining information from the text and structure embeddings to learn better representation than using a joint loss function. 

From the ablation studies, one can conclude that our novel context-aware architecture and adversarial training technique helps the model ACNE in achieving state-of-the-art performance in link prediction tasks.

%%%%%%%%%%%%%%%%%%%%%%%%%%%%%%%%%%%%%%%%%%%%%%%%%%% 
\begin{table}[!ht] 
\begin{center}
 \begin{tabular}{|c | c| c | c | c | c |c | c| c| c}
 \hline
 $\%$ of Nodes & 10$\%$ & 30$\%$ & 50 $\%$ & 70$\%$  \\  
 \hline
%LINE$^a$ & 53.9 & 56.7 & 58.8 & 60.1 \\
%TADW$^b$  & 71.0 & 71.4 & 75.9 & 77.2 \\
CANE$^a$ & 81.6 &	82.8 &	85.2	& 86.3 \\
DMTE$^b$ & 81.8 & 	83.9 & 	86.3 &  87.9	 \\
WANE$^c$ & 81.9 & 83.9 & 86.4 & 88.1 \\
GANE$^d$ & 82.3 & 84.2 & 86.7 & 88.5  \\
VHE$^e$  &  82.6 & 84.3 & \textbf{87.7} & 88.5 \\
\hline 
CNE-top (baseline) &  76.9 & 81.6 & 83.5 &  84.3  \\
CNE-mu (baseline) & 82.5 & 84.5 & 86.0 & 88.1\\
CNE (baseline) & 83.1 & 85.7  & 87.0 & 88.4 \\
%MCNE & 83.1 & 85.8 & 87.2 & 88.0 \\
\hline 
ACNE(ours)  & \textbf{83.5} & \textbf{86.2} & \textbf{87.7} & \textbf{88.7} \\
%AMCNE(ours) & 83.5 & 86.5 & 87.5 & 88.5 \\
\hline
\end{tabular}
\caption{F1-Marco score for node classification in Cora. Here a higher value of F1-score indicates better performance. \textbf{Citations:} $^a$\cite{tu-etal-2017-cane}, $^b$\cite{Zhang:2018:DMTE}, $^c$\cite{shen-etal-2018-improved}, $^d$\cite{GANE2019}, $^e$\cite{wang2019improving}   }
\label{tab:classcora}
\end{center}
\end{table}

\subsection{Unseen Nodes Link Prediction}
For predicting structure embeddings of unseen nodes, we use the post-training in Section \ref{sec: unseennodes}. The results of this experiment are shown in Table \ref{tab:linkpredunseen}. We can see that our model ACNE give better performance than the state-of-the-art model VHE in all settings. The results for other models are taken from paper \cite{wang2019improving}. We have also created a baseline model using the mapping function on the CNE model. Unlike models CANE, WANE and CNE our model have unsupervised post-training state. This will improve the performance considerably by using the text-similarity to provide supervision. As a result, we can see that our model considerably outperforms these models in the low-data regime. In model VHE, they use a variational autoencoder framework with homophilic-prior to learn embeddings for unseen nodes. But our model ACNE outperforms it because our model uses text similarity directly whereas VHE uses it through an autoencoder framework. We also can see in Cora and Hepth, our baseline model CNE performs poorly in the low-data regime and gets better results in text-matching based models CANE and WANE when training size is larger. This because the mapping function learns better map from text to structure embeddings, so the quality of topological attention in CNE improves. In Zhihu dataset, we can see similar performance improvement as observed in the link prediction performance of seen nodes. This because of Zhihu being a larger network and topological attention we use in the models CNE and ACNE.

%%%%%%%%%%%%%%%%%%%%%%%%%%%%%%%%%%%%%%%%%%%%%%%%%%%%%%%%%%%%%%%%%%%%%%%%%%%%%%

\subsection{Node Classification}
The goal of this task is to assign a label to each of the nodes. This is done by using the node embeddings, after learning on the entire graph, as features for training a linear-SVM classifier. We evaluated the model on different levels of supervision by dividing the embeddings into train and test by varying the percentage of labelled data used from $10\%$ to $70\%$ in steps of $20$. We used Cora dataset that has supervised node-labels for this task. The SVM model from Sklearn \cite{scikitlearn2011} with default parameter settings was used for training the classifier. The results for our model is shown in Table \ref{tab:classcora}. We can see that our model ACNE give considerable improvement over previous state-of-the-art VHE in all settings. As we explained earlier, adversarial treatment improves performance that can be observed as the model ACNE performs better than its non-adversarial counterpart CNE. In addition to that, CNE performs better than CNE-mu and CNE-top, this because using both attention mechanism helped the model to learn better representations. We can also see that CNE gives performance comparable to or even beats the attention-based text-matching models like CANE, WANE,  and GANE. So, we can infer that our novel architecture and adversarial training of the model helped us in getting state-of-the-art performance in this task.

%%%%%%%%%%%%%%%%%%%%%%%%%%%%%%%%%%%%%%%%%%%%%%%%%%%%%%%%%%%%%%%%%%%%%%%%%%%%%%%%%

\section{Related Work}
 The techniques for textual network representation can be categorized into two kinds. The first uses fixed embeddings for representing textual information. These include models like TADW \cite{Yang:2015:NRL:TADW}, which uses matrix factorization to learn text embeddings, and models like  Tri-DNR \cite{Pan:2016:TriDN}, CENE \cite{CENE2016}, and DMTE \cite{Zhang:2018:DMTE}, which uses a skip-gram model to learn text embeddings. 

The second kind of models use context-aware embeddings. Recent state-of-the-art models are on this line of work. CANE \cite{tu-etal-2017-cane} uses a mutual attention mechanism to learn affinity/similarity matrix between text features across a node pair. That is then used for finding context-aware embeddings by  weighted average of the text features. WANE \cite{shen-etal-2018-improved} uses a mutual attention mechanism for aligning word embeddings of the textual content in one node to the other. Then pooling is applied to find the final embedding.  GANE \cite{GANE2019} uses optimal transport framework for attention instead of dot product for finding the affinity matrix between words in a node-pair. NEIFA \cite{xu-etal-2019-deep} uses gating mechanism for getting context-aware embedding and also it uses gating for extracting information from structure embedding that is complementary to textual information. VHE \cite{wang2019improving} uses a variational framework for learning context-aware embedding using a mutual attention mechanism. 

Apart from previously discussed methods on textual networks, GraphGAN \cite{graphgan2018} used for representation learning in homogeneous networks uses a similar adversarial framework as ours by using generator and discriminator networks. Here, both generator and discriminator separately learn embeddings for a node, but this model uses only the structural information. So, the performance of this model will be poor in the case of representation learning of sparse textual networks.

\section{Conclusions}
In this paper, we propose a new technique called Adversarial Context-Aware Network embeddings (ACNE) for representation learning in textual networks. This method uses an adversarial framework between text embeddings and structure embeddings for modality fusion.
%In the proposed approach, embeddings of the nodes are the result of an adversarial game between a discriminator and a generator. The generator is based on structure embedding to model the connectivity of the network, and the discriminator uses the text encoding for predicting edges between a pair of nodes. For discriminator, we have developed a new text-encoding method that combines both mutual and topological attention to learn text embeddings. 
Through extensive experiments, we have shown that our model achieves state-of-the-art results in the task of link prediction and node classification. We also demonstrated the effectiveness of topological attention in improving the performance of the context-aware network embeddings. We have shown that one of the main advantage of the proposed model is that it can be extended for learning unseen nodes in training using a post-training stage. This is achieved by using discriminator for providing supervision for learning structure embeddings in the generator. We also achieved state-of-the-art results in the link prediction task on the unseen nodes.

\bibliography{final}

\end{document}